\pdfoutput=1
\documentclass{article} 
\usepackage{nips14submit_e,times}
\usepackage{url}
\usepackage{graphicx}
\usepackage{amssymb}
\usepackage{amsmath}
\usepackage{tabularx}
\usepackage{multirow}
\usepackage{booktabs}

\usepackage{color}

\def\X{{\bf X}}
\def\x{{\bf x}}
\def\Y{{\bf y}}

\def\W{{\bf W}}
\def\w{{\bf w}}

\def\eg{\textit{e.g.,~}}
\def\cf{\textit{cf.~}}
\def\ie{\textit{i.e.~}}

\def\etal{\textit{et~al.~}}
\def\iid{\textit{i.i.d.~}}

\usepackage[font=small,labelfont=bf,tableposition=top]{caption}
\usepackage[font=footnotesize]{subcaption}
\usepackage{algorithmic}
\usepackage{algorithm,algorithmic}
\definecolor{lgray}{RGB}{225,225,225}
\definecolor{mgray}{RGB}{185,185,185}
\definecolor{lightgray}{gray}{0.40}
\newcommand{\algcomment}[1]{{\color{lightgray} \# #1}}

\usepackage[pagebackref=true,breaklinks=true,letterpaper=true,colorlinks,bookmarks=false]{hyperref}

\title{Self-Learning Camera: Autonomous Adaptation of Object Detectors to Unlabeled Video Streams}

\author{
Adrien~Gaidon\thanks{adrien.gaidon@xrce.xerox.com} \\
Computer Vision group\\
Xerox Research Center Europe, France \\
\texttt{} \\
\And
Gloria~Zen \\
DISI, University of Trento, Italy \\
\AND
Jose  A.~Rodriguez-Serrano\\
Machine Learning for Services group\\
Xerox Research Center Europe, France \\
}

\nipsfinalcopy 

\begin{document}

\maketitle

\begin{abstract}
Learning object detectors requires massive amounts of labeled training samples
from the specific data source of interest.
This is impractical when dealing with many different sources (\eg in camera
networks), or constantly changing ones such as mobile cameras (\eg in
robotics or driving assistant systems).
In this paper, we address the problem of \emph{self-learning} detectors in an
\emph{autonomous} manner, \ie (i) detectors continuously
updating themselves to efficiently adapt to streaming data sources (contrary to
transductive algorithms), (ii) without any labeled data strongly related to the
target data stream (contrary to self-paced learning), and (iii) without manual
intervention to set and update hyper-parameters.
To that end, we propose an unsupervised, on-line, and self-tuning learning
algorithm to optimize a multi-task learning convex objective.
Our method uses \emph{confident but laconic oracles} (high-precision but low-recall
off-the-shelf generic detectors), and exploits the structure of the problem to
jointly learn on-line an ensemble of instance-level trackers, from which we
derive an adapted category-level object detector.
Our approach is validated on real-world publicly available video object
datasets.
\end{abstract}

\vspace*{-3mm}
\section{Introduction}
\vspace*{-1mm}

The theoretical and practical success of many machine learning algorithms
often rests on two fundamental assumptions: (i) access to many independent and
identically distributed (\iid) samples, (ii) stationarity of the underlying data
distribution (and in particular, that it does not change from the observed samples
to the unobserved ones).
This explains why data collection and labeling are crucial tasks for most
supervised learning algorithms. In practice, these tasks are particularly
challenging in videos, due in part to the large volume and variability
typically observed in video collections.
%

In this paper, we focus on the problem of on-line unsupervised learning of
classifiers, where the data source is a stream of unlabeled and non-\iid samples
drawn from a non-stationary distribution.
In particular, we propose an unsupervised adaptation method to learn models
(video object detectors) that automatically and continuously adapt over
time, which we refer to as \emph{self-learning} and \emph{autonomous
adaptation}.

\begin{figure}
    \vspace*{-4mm}
    \includegraphics[width=\linewidth]{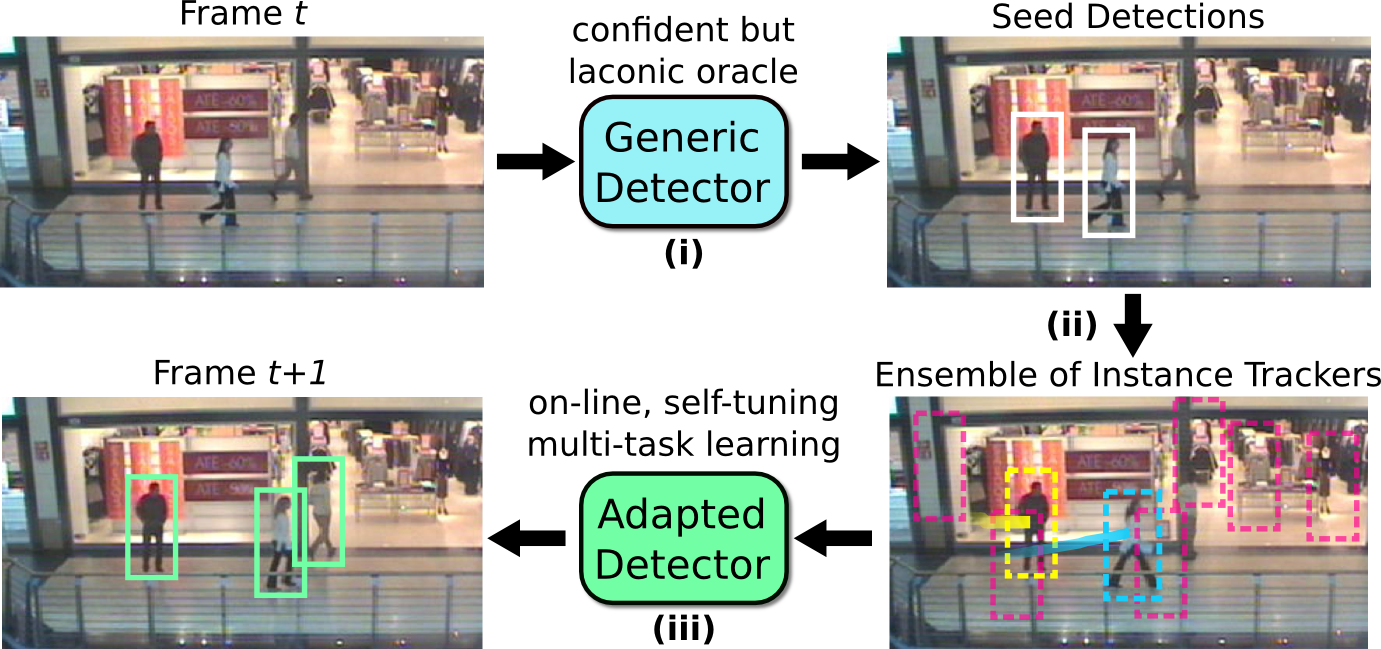}
    \vspace*{-2mm}
    \caption{
        \label{fig:overview}
        Overview of our approach.
        \textbf{(i)} A \emph{confident but laconic oracle} is periodically run
        along an unseen video stream to generate a few \emph{seed} detections
        (candidate bounding boxes with top confidence scores).
        \textbf{(ii)} The seed objects are tracked along the video with an
        \textbf{Ensemble of Instance Trackers (EIT)}. \emph{Our main contribution is
        that we jointly learn the detectors used by each tracker using an on-line
        self-tuning multi-task learning algorithm} able to leverage \emph{difficult}
        samples (in purple).
        %
        %
        \textbf{(iii)} An adapted model is available anytime from the EIT, and
        allows to detect objects --- beyond just the few seed objects --- in the rest
        of the stream or other videos.
    }
    \vspace*{-2mm}
\end{figure}

Instead of relying on labeled data strongly related to the target distribution,
we assume only the availability of \emph{confident but laconic oracles}:
black-box classifiers pre-trained on standard general-purpose datasets in a
high-precision but low-recall regime, so that they can automatically label a
few of the easiest target samples.
We propose to learn a classifier from the underlying latent commonalities
between these samples using an on-line multi-task learning formulation allowing
to automatically label and incorporate more difficult target
samples (in particular negative ones) along a data stream.
Consequently, our approach differs from most unsupervised domain adaptation
methods, as we do not rely only on easy target samples (in contrast to works
inspired by the self-paced learning of Kumar~\etal~\cite{Kumar2010}), and we
are not limited to the stationary and transductive learning settings (see
Section~\ref{s:relwork} for a more detailed discussion).
%
%
Figure~\ref{fig:overview} presents an overview of our approach.

Self-adapting a classifier might suffer from negative transfer, either because
of over-fitting to only a few positive samples, or because of drifting due to the
inclusion of false positives. Another challenge is that there is in general no
way to tune hyper-parameters due to the lack of labeled target validation data.
In this work, we address these issues by leveraging the spatio-temporal
structure of video data.
%
First, in Section~\ref{s:mtl}, we propose to use multi-task learning in
conjunction with tracking as a way to automatically elicit a useful set of
target positives and negatives, especially ``hard negatives'', which are
crucial for high detection performance~\cite{Dalal:05,DPM}.
Our multi-task learning objective yields an \emph{adapted shared latent model of
the category}, which bridges the gap between instance-level models used
for tracking and category-level models used for detection.
%
Second, in Section~\ref{s:selftuning}, we detail how to learn this model. We
propose a novel efficient on-line algorithm applicable to non-stationary
streaming scenarios. Our method rests on the combination of recent stochastic
optimization techniques with a novel hyper-parameter self-tuning method
exploiting a ``no-teleportation and no-cloning'' assumption.
Experiments, in Section~\ref{s:experiments}, on two challenging real-world
datasets illustrate that we are able to learn and adapt detectors from scratch
on different scenes with no labeled data.

\section{Related work}
\label{s:relwork}


Unsupervised domain adaptation approaches use annotated data from a fixed
dataset (the source set) and only \emph{unlabeled} data from the target dataset.
They often require one or more passes over a large pool of annotated samples
in order to adapt the model to each new target dataset, which makes them
impractical for continuous adaptation to streaming data.
For instance, Taskar~\etal~\cite{Taskar2003} leverage ``unseen'' features from
instances classified with high confidence, learning which features are useful
for classification,
%
%
and Gong~\etal~\cite{Gong2013} ``reshape'' datasets to minimize their
distribution mismatch.
%
%
These methods assume that the target domain is stationary, with many unlabeled
target samples readily available at each update (transductive
setting~\cite{Joachims1999}).
Consequently, these approaches are not suited to non-stationary streaming
scenarios, such as when using on-board cameras (\eg for autonomous driving or
robotics).

Most related to our work are methods exploiting the spatio-temporal structure
of videos in order to collect target data samples.
Prest~\etal~\cite{Prest2012} learn object detectors by relying on joint motion
segmentation of a set of videos containing the object of interest moving
differently from the background. Their goal is to learn generic detectors that
adapt well from video to image data.
%
Tang~\etal~\cite{Tang:12} are inspired by the self-paced learning approach of
Kumar~\etal~\cite{Kumar2010}, \ie ``learn easy things first'', which is
designed for fine-tuning on target domains related to a labeled source one.
The transductive approach of Tang~\etal~\cite{Tang:12} iteratively re-weights
labeled source samples by using the tracks of the closest target samples.
%
Sharma~\etal~\cite{Sharma2012} propose a similarly inspired multiple instance
learning algorithm, also relying on self-paced learning and off-line iterative
re-training of a generic detector.
These approaches can only be applied in stationary transductive settings, and
do not allow for efficient model adaptation along a particular video stream.

Several tracking-by-detection methods are also related to our work, in
particular the tracking-learning-detection approach of Kalal~\etal~\cite{PN}
and the multi-task learning approach of Zhang~\etal~\cite{zhang2013robust}.
However, they are designed to work on short sequences, and their goal is to
learn only instance-specific models, whereas our aim is to learn and adapt
\emph{category-level} ones.
Note that we also differ from the family of multi-class transfer learning
methods aiming to share features across categories, such as the ``learning to
borrow'' approach of Lim~\etal~\cite{Lim2011}.
%
%
These approaches are different in intent from our continuous category-level
adaptation to a new data stream.  Furthermore, they are not straightforwardly
applicable for object detection with one or few categories of interest (\eg
cars and pedestrians).

\section{Multi-task learning of an ensemble of trackers}
\label{s:mtl}

\subsection{Tracking-by-detection}

In this work, the term ``object detector'' denotes a binary classifier
parametrized by a (learned) vector $\w \in \mathbb{R}^d$. This classifier
computes the probability that a window --- an image sub-region $\x$
represented by a feature vector $\phi(\x) \in \mathbb{R}^d$ --- contains
an object of the category of interest by:
\begin{equation}
P(\x) = \left(1 + e^{-\left( \w^T \phi(\x) + b \right)}\right)^{-1}.
\label{eq:lr}
\end{equation}
%
%

When a new frame is available, the first step of our method consists in calling
a \emph{confident but laconic oracle} returning a few (or no) automatically labeled
samples (image regions with object category labels in our case). 
We assume that the automatic labels of these seed target samples are mostly
correct, which means that they correspond only to the ``easiest'' target samples
(high-precision but low-recall regime).
%
%
%
%
We track each of these seeds in subsequent frames using a simple but efficient
tracking-by-detection algorithm inspired by the P-N learning algorithm of
Kalal~\etal~\cite{PN}.
The tracker associated to each seed $i$ maintains its own personal detector
$\w_i^{(t)}$.  This detector is run on a new frame $t+1$ to return a pool of
candidate locations.
The new location of object $i$ is then simply the candidate region that yields
the closest match with the location in the previous frame (after simple motion
interpolation based on sparse optical flow).
%
If there is no match, then the object $i$ is considered lost, and it is not
tracked further.
If a new location is found, then the corresponding region is considered as
positive and the other detections as negatives in order to update the model
$\w_i^{(t)}$ of object $i$.
This relies on both a spatio-temporal smoothness assumption (based on the
observed motion of each object instance), and the uniqueness of each object
being tracked. In other words, we assume there is neither teleportation, nor
cloning of the individual instances.

Note that these negatives are in fact \emph{false positives}, \ie ``difficult''
target samples, also called \emph{hard negatives}~\cite{Dalal:05}. Contrary to
the easier samples used in self-paced learning (\eg in
Tang~\etal~\cite{Tang:12}), they are key to the successful update of the model.
This is, indeed, well established in the object detection community (\cf
Felzenszwalb~\etal~\cite{DPM} for instance). Furthermore, we conducted
preliminary experiments where directly learning a video-specific detector on
seed samples as positives and random negatives consistently resulted in
negative transfer (\ie worse performance than the generic detector oracle
alone).
We also observed this negative transfer when replacing the automatic seed
positives by the \emph{ground truth} bounding boxes of the moving objects
only.
Note that our tracking approach does not rely on background subtraction or
motion segmentation, and is therefore applicable to stationary objects and
moving cameras.

This negative transfer illustrates the difficulty of bridging the gap between
instance-level models and category-level ones, and the importance of difficult
target samples. Note that there is also a risk of negative transfer when using
difficult samples (as they might be wrongly labeled).
In the following, we describe how we address these challenges via our novel
multi-task tracking formulation to exploit the commonalities between seeds.

\subsection{The Ensemble of Instance Trackers (EIT) model}

Let $N_{t+1}$ be the number of object instances currently being tracked, each
one associated with its own detector $\w_i^{(t)}$.
Updating each detector amounts to minimizing the regularized empirical risk: 
\begin{equation}
\w_i^{(t+1)} = \arg\min_{\w_i} \sum_{k=1}^{n_i} \ell(\x_k^{(i)},y_k^{(i)},\w_i) + \lambda\Omega(\w_i),
\quad i=1, \cdots, N_{t+1}
\label{eq:riski}
\end{equation}
where $\x_k^{(i)}$ is one of the $n_i$ training samples of object $i$ in frame
$t+1$ --- either a random window (to get negatives during initialization), a seed
from the generic detector, or a candidate detection obtained by running the
current version of the model $\w_i^{(t)}$ --- and $y_k^{(i)}$ is a binary
label. The labels are obtained using the aforementioned ``no teleportation and
no cloning'' assumption, with the addition that negative samples overlapping any
other object currently tracked are not used (a region that is not Ann's car
does not mean it is not a car).
Here, $\ell(\x,y,\w)$ is the loss incurred by classifying $\x$ as $y$ using
parameters $\w$, and $\Omega(\w)$ is a regularizer.
In our experiments, we use the logistic loss:
\begin{equation}
\ell(\x, y, \w) = 
\log \left( 1 + \exp \left(-y \left( \w^T \phi(\x) + b \right)\right)\right) \ ,
\end{equation}
as this gives calibrated probabilities with Eq.~\eqref{eq:lr}, and enjoys useful
theoretical properties for on-line optimization~\cite{Bach2013}.
The set of $N_{t+1}$ trackers, with joint detector parameters
$\W^{(t)}=\{\w_1^{(t)},\ldots,\w_{N_{t+1}}^{(t)}\}$ is denoted as \textit{Ensemble of Instance
Trackers} (EIT). With this notation, we can express the $N_{t+1}$ equations in
Eq.~\eqref{eq:riski} as a joint minimization over the ensemble parameters:
\begin{equation}
\W^{(t + 1)} = \arg\min_{\W} L(\X^{(t + 1)},\Y^{(t + 1)},\W) + \lambda \Omega_t(\W),
\label{eq:risk}
\end{equation}

Clearly, if $L(\X^{(t + 1)},\Y^{(t + 1)},\W)=\sum_i \sum_k \ell(\x_k^{(i)},y_k^{(i)},\w_i)$ and
$\Omega_t(\W)=\sum_i \Omega(\w_i)$, we recover exactly Eq.~\eqref{eq:riski}
where each classifier is learned independently. 
In order to jointly learn all the classifiers, we impose the following
multi-task regularization term instead: 
\begin{equation}
\label{eq:reg}
\Omega_t(\W) = \frac{1}{2N_{t+1}}\sum_{i=1}^{N_{t+1}} \|\w_i - \bar{\w}^{(t)} \|_2^2. 
\end{equation}
where $\|\w_i\|_2$ denotes the $l_2$ norm of $\w_i$, and $\bar{\w}^{(t)}$ is
the (running) mean of all previous instance models, which comprises all past
values of the models of currently tracked or now lost seed-specific detectors.
Note that this formulation is closely related to the mean-regularized
multi-task learning formulation of Evgeniou and Pontil~\cite{Evgeniou:04}, with
the difference that it is designed for on-line learning in streaming scenarios.
This regularization promotes solutions where classifiers (past and present) are
similar to each other (in the Euclidean sense).
Therefore, this regularization prevents each detector from over-fitting to the
appearance of the individual object appearances, and allows them to generalize
across tasks (object instances) by modeling the latent commonalities (the
appearance of the category) with the mean of the trackers.

Importantly, this provides a theoretical justification to using the running average
$\bar{\w}^{(t)}$ as a single category-level detector.
Consequently, once the detectors $\w_i$ are updated in frame $t+1$, a new
scene-adapted detector is readily available as:
\begin{equation}
\bar{\w}^{(t + 1)} = 
\frac{1}{\bar{N}_t + N_{t+1}}
\left(
\bar{N}_t \bar{\w}^{(t)}
+
\sum_{i=1}^{N_{t+1}} \w_i^{(t+1)}
\right) \ ,
\label{eq:final}
\end{equation}
where $\bar{N}_t = \sum_{\tau = 1}^t N_{\tau}$.
%
%
In our object detection case, this approach can be interpreted as learning a
category-level model from the average of instance-specialized models. As we use
linear classifiers, this multi-task learning could be seen as an improvement of
``late fusion'' of exemplar-based models, such as the Exemplar-SVM of
Malisiewicz~\etal~\cite{exemplar}. A major difference is that our models are
learned \emph{jointly} and \emph{adapt continuously} to both the data stream
and other exemplars.

Another benefit of this regularization is that it learns a model
that is more robust to erroneous seeds (initial false detections of the generic
detector), as they are likely to significantly differ from the mean, and,
therefore, the corresponding trackers will be quickly under-fitting and lose
the object.  In contrast, the correct seeds will be tracked for longer, as they
share common appearance factors, thus contributing more to the category model.
Note, however, that the $l_2$-norm used in Eq.~\eqref{eq:reg} is not robust to
outliers. Therefore, the handling of incorrect seeds could be improved by using
more robust alternatives --- \eg sparsity-inducing or trace-norm regularizers,
replacing the mean by the median, or outlier detection methods --- at the cost
of a more computationally demanding solution, which is also more complex to
update on-line than our running average.
Our approach is compatible with these improved regularizers, but we focus on
the simple regularization of Eq.~\eqref{eq:reg} in order to directly show the
potential of our self-tuning on-line multi-task learning algorithm described
next.


\section{Continuous self-tuning on-line adaptation}
\label{s:selftuning}

\subsection{Streaming asynchronous optimization}

We solve the optimization problem in Eq.~\eqref{eq:risk} using Averaged
Stochastic Gradient Descent (ASGD), also called Polyak-Rupert
averaging~\cite{ASGD}. This on-line first order method achieves optimal
convergence rate with only one pass over the data and a constant learning rate
with a logistic regression model~\cite{Bach2013}.
The update rule for each model $\w_i$ is:
%
\begin{equation}
\w_i^{(t+1, k)} = \w_i^{(t+1, k - 1)} -\eta \left( \frac{\partial \ell}{\partial \w}
(\x_k^{(i)}, y_k^{(i)}, \w_i^{(t+1, k-1)}) +
\frac{\lambda}{N_{t+1}} \left(\w_i - \bar{\w}^{(t)} \right) \right),
\label{eq:sgd}
\end{equation}
where $\eta$ is the learning rate, $(\x_k^{(i)}, y_k^{(i)})$ a
training sample, $k=1,\cdots,n_i$, and $\w_i^{(t+1, 0)} = \w_i^{(t)}$.
The actual model used is then the average of all previous updates, which is
simply obtained by maintaining a running mean of the updates to compute:
\begin{equation}
\w_i^{(t + 1)} = \frac{1}{n_i} \sum_{k=1}^{n_i} \w_i^{(t+1,k)}
\label{eq:instanceupdate}
\end{equation}
Equations~\eqref{eq:risk},\eqref{eq:reg}, \eqref{eq:final} and \eqref{eq:sgd}
show that the learning process is a joint one: the update of the detector
$\w_i$ includes a contribution of all the other detectors (both current and
past ones).
Another advantage is that our approach is well suited to a streaming and
asynchronous scenario, as the update of each specific model $\w_i$ only relies
on the current frame and the current running average of all models.
In practice, we use a mini-batch variant of ASGD with potentially multiple
passes over the data available at time $t+1$. This allows to share window
preprocessing operations at the frame level, and reduces the variance
of the gradients.
We also maintain the running average of the parameters not only across gradient
descent steps, but also across frames (not depicted in
Eq.~\eqref{eq:instanceupdate} for simplicity).

\subsection{Self-tuning of hyper-parameters}

A key issue in our unsupervised on-line learning setting is how to select the
values of the different hyper-parameters involved, amongst which are the
learning rate $\eta$ in Eq.~\eqref{eq:sgd}, the regularization parameter $\lambda$
in Eq.~\eqref{eq:risk}, and the number of iterations per mini-batch.
This is particularly important, as we have to continuously update appearance
models in a streaming fashion, where data samples are likely to be non-\iid,
and the video stream to be non-stationary (\eg in long-term surveillance
scenarios).
The learning rate, in particular, is a crucial parameter of ASGD, and should be
set on a per-update basis in this non-stationary case.

Cross-validation, typically used to tune hyper-parameters, is not
applicable in our setting, as we have only one positive example at a time.
Furthermore, optimizing the window classification performance is not guaranteed
to result in optimal detection performance due to the additional
post-processing steps applied in most detection methods~\cite{Cinbis:13}.

\begin{figure}[!ht]
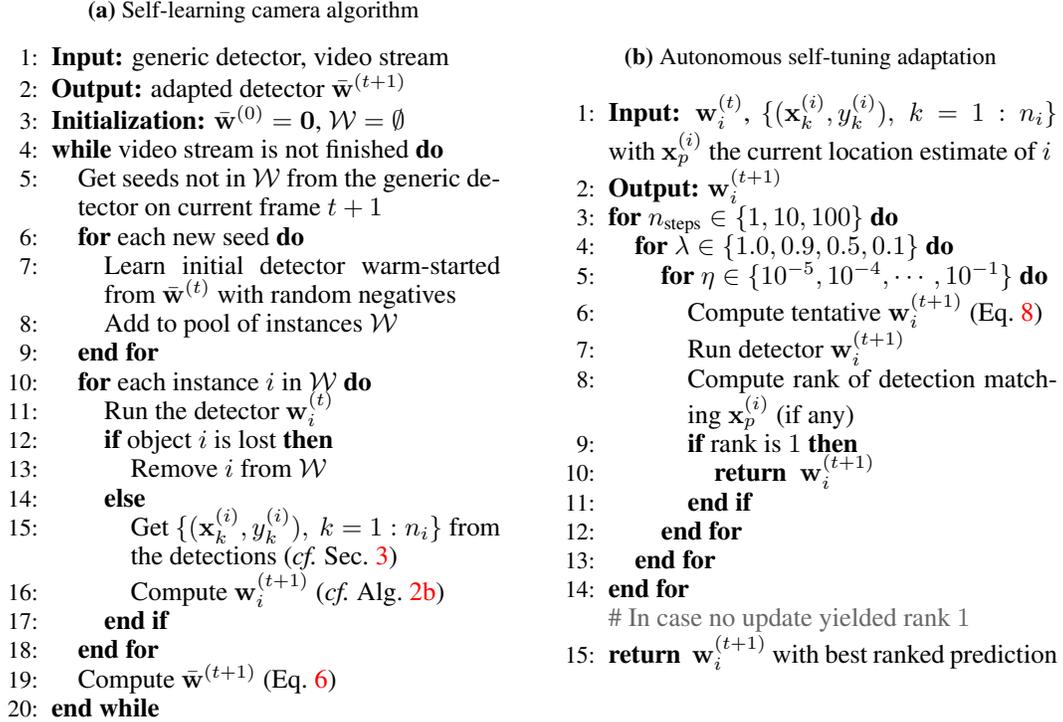

\centering
\begin{subfigure}{.47\textwidth}
\centering
\caption{Self-learning camera algorithm}\label{algo:slc}
\begin{algorithmic}[1]
    \STATE \textbf{Input:}
        generic detector, video stream
    \STATE \textbf{Output:}
        adapted detector $\bar{\w}^{(t+1)}$
    \STATE \textbf{Initialization:} $\bar{\w}^{(0)} = \mathbf{0}$, $\mathcal{W} =
           \emptyset$ 
    \WHILE{video stream is not finished}
    \STATE Get seeds not in $\mathcal{W}$ from the generic detector on current
           frame $t + 1$
    \FOR{each new seed}
        \STATE Learn initial detector warm-started from
           $\bar{\w}^{(t)}$ with random negatives
        \STATE Add to pool of instances $\mathcal{W}$
    \ENDFOR
    \FOR{each instance $i$ in $\mathcal{W}$}
        \STATE Run the detector $\w_i^{(t)}$
        \IF{object $i$ is lost}
            \STATE Remove $i$ from $\mathcal{W}$
        \ELSE
            \STATE Get $\{(\x_k^{(i)}, y_k^{(i)}), \ k=1:n_i \}$
                   from the detections (\cf Sec.~\ref{s:mtl})
            \STATE Compute $\w_i^{(t+1)}$ (\cf Alg.~\ref{algo:selftuning})
        \ENDIF
    \ENDFOR
    \STATE Compute $\bar{\w}^{(t+1)}$ (Eq.~\ref{eq:final})
    \ENDWHILE
\end{algorithmic}
\end{subfigure}
    \hfill
\begin{subfigure}{.47\textwidth}
\centering
\caption{Autonomous self-tuning adaptation}\label{algo:selftuning}
\begin{algorithmic}[1]
    \STATE \textbf{Input:}
        $\w_i^{(t)}$, $\{(\x_k^{(i)}, y_k^{(i)}), \ k=1:n_i \}$
        with $\x_p^{(i)}$ the current location estimate of $i$
    \STATE \textbf{Output:}
        $\w_i^{(t+1)}$
    \FOR{$n_{\text{steps}} \in \{1, 10, 100\}$}
    \FOR{$\lambda \in \{1.0, 0.9, 0.5, 0.1\}$}
    \FOR{$\eta \in \{10^{-5}, 10^{-4}, \cdots, 10^{-1}\}$}
        \STATE Compute tentative $\w_i^{(t+1)}$  (Eq.~\ref{eq:instanceupdate})
        \STATE Run detector $\w_i^{(t+1)}$ 
        \STATE Compute rank of detection matching $\x_p^{(i)}$ (if any)
        \IF{rank is $1$}
            \RETURN $\w_i^{(t+1)}$
        \ENDIF
    \ENDFOR
    \ENDFOR
    \ENDFOR
    \\ \algcomment{In case no update yielded rank $1$}
    \RETURN $\w_i^{(t+1)}$ with best ranked prediction
\end{algorithmic}
\end{subfigure}
\caption{Pseudo-code overview of our approach (book-keeping operations like
maintaining counters and running averages are not described for simplicity,
refer to the main text for more details)}\label{fig:pseudocode}
\end{figure}

Instead, we use a strategy consisting in greedily searching for the
least-over-fitting parameters --- \eg smallest $\eta$ and number
of iterations, largest $\lambda$ --- that optimize the rank of
the correct detection in the current frame.
This only involves running the detector after each tentative update in order to
assess whether it yields a top ranked detection strongly overlapping with the
tracker's prediction, and is efficient in practice, as it re-uses the already
extracted frame-level computations (in particular the feature extraction).
See Figure~\ref{fig:pseudocode} for a high-level pseudo-code description of our
method.


\section{Experiments}
\label{s:experiments}

\subsection{Datasets}

\begin{figure}
\center
\includegraphics[height=4.0cm]{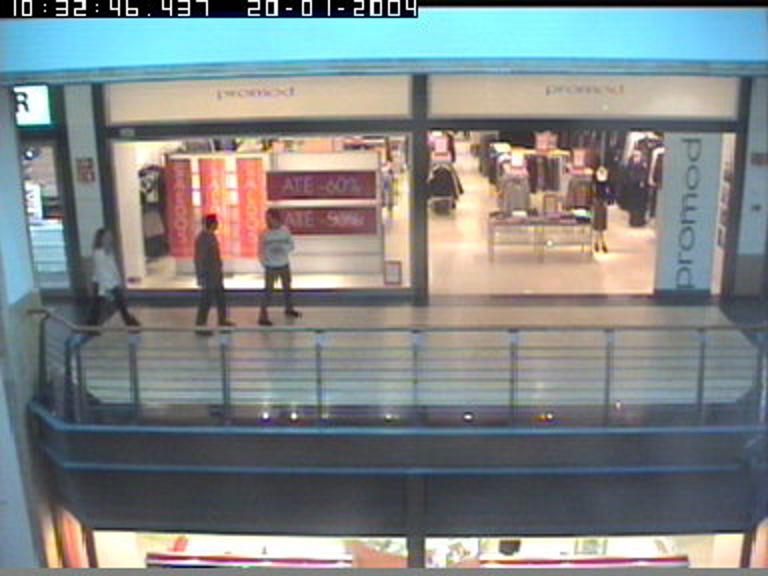} \
\includegraphics[height=4.0cm]{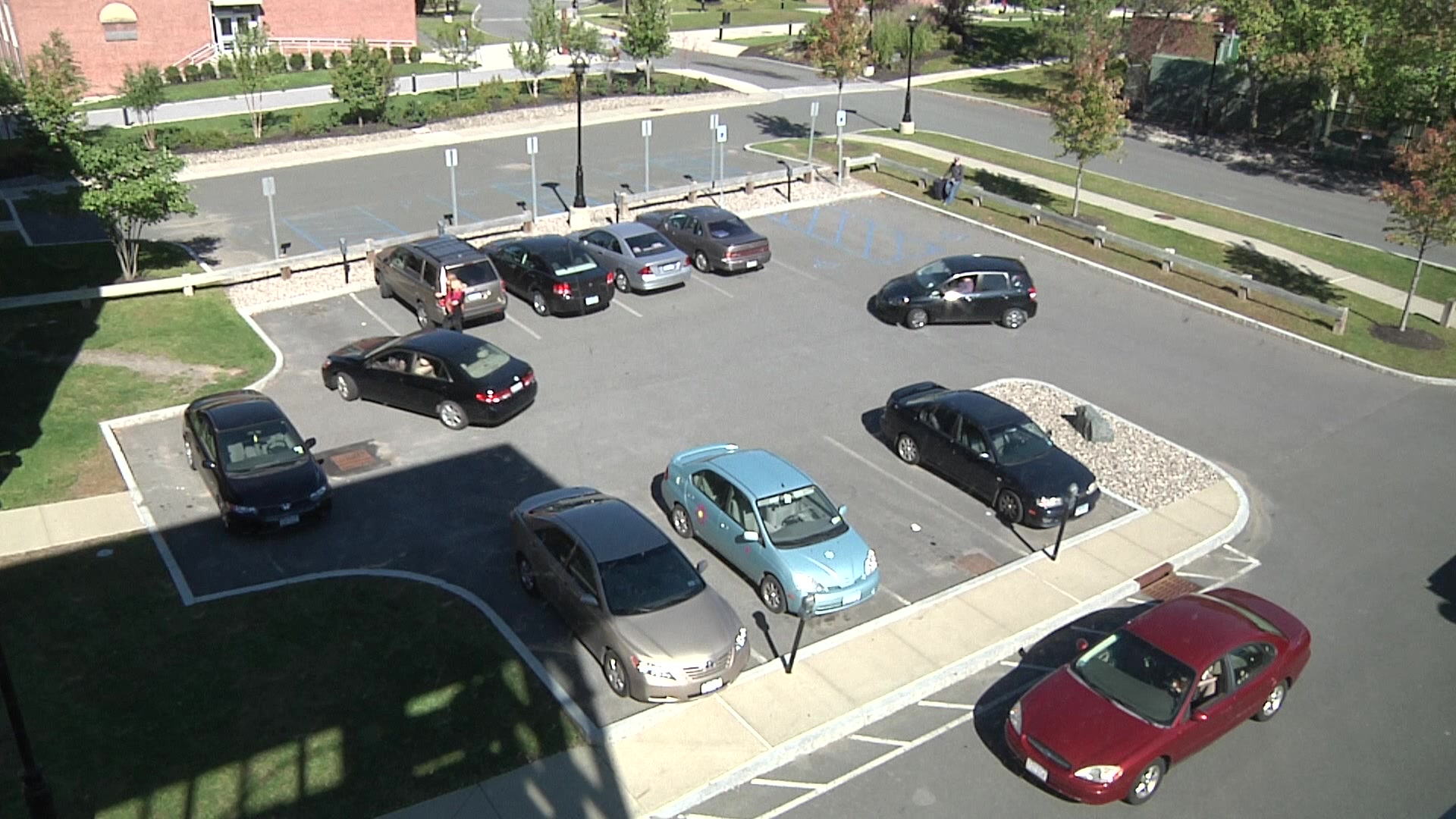}
\caption{\label{fig:data} Frames from the CAVIAR (left) and VIRAT-0401 (right) datasets.}
\end{figure}

In order to validate our approach, we use publicly available benchmarks: five
video sequences from
CAVIAR\footnote{http://homepages.inf.ed.ac.uk/rbf/CAVIAR/}, and a long HD video
sequence from VIRAT\footnote{http://www.viratdata.org}.
Objects of interest in CAVIAR are persons, while VIRAT objects of interest
are cars.
Sample images from the two scenarios are shown in Figure~\ref{fig:data}.
Some dataset details are reported in Table~\ref{tab:data}.

On CAVIAR, we use the same sequences as Wang~\etal~\cite{WHG:12} --- Ols1,
Ols2, Osow1, Olsr2, and Ose --- and also upscale the image size by a factor of
2.0. We run independent experiments on each of the five sequences.
Each sequence is used both for unsupervised learning and evaluation (\ie the
same setting as in \cite{WHG:12}). In our case, this corresponds to a scenario
where the object detector is continuously adapted along a finite video stream,
to later be applied to the entirety of this video.
Note that we compare to Wang~\etal~\cite{WHG:12}, a state-of-the-art
unsupervised video adaptation method also relying only on a pre-trained generic
detector, and assuming the most confident detections are true positives. In
contrast to our approach however, Wang~\etal~\cite{WHG:12} propose a ``lazy
learning'' method consisting in re-ranking the low scoring detections using
the similarity to high scoring ones.  While this re-ranking process improves
the detection performance, the method does not produce an adapted detector, and
can, therefore, only be applied in a batch setting, as on the CAVIAR dataset.

For VIRAT, we use sequence 0401, because it is fully annotated with both
static and moving objects, and corresponds to a typical parking scene.
The size (375K objects) and duration of VIRAT-0401 (58K frames) allows
splitting the sequence in two parts (of equal length). Our self-learning
algorithm is run on the first part of the video to self-learn and autonomously
adapt the specific detector, whose generalization performance (on different
scenes from the same target domain) is then evaluated on the second part of the
video stream.

\begin{table}
\begin{center}
\begin{tabular}{lccccc}
               & frame size         & fps &  \#frames & class  & \#objects
\\
\toprule
CAVIAR (Ols1)  & 576 $\times$ 768   & 25  & 295   & pedestrian  & 438    \\
CAVIAR (Ols2)  & 576 $\times$ 768   & 25  & 1119  & pedestrian  & 290    \\
CAVIAR (Osow1) & 576 $\times$ 768   & 25  & 1377  & pedestrian  & 2402   \\
CAVIAR (Olsr2) & 576 $\times$ 768   & 25  & 560   & pedestrian  & 811    \\
CAVIAR (Ose2)  & 576 $\times$ 768   & 25  & 2725  & pedestrian  & 1737   \\
VIRAT-0401     & 1080 $\times$ 1920 & 30  & 58$K$ & car         & 375$K$ \\
\bottomrule
\end{tabular}
\end{center}
\caption{Video object detection datasets used in our experiments}
\label{tab:data}
\vspace*{-5mm}
\end{table}

\subsection{Implementation details}

\subsubsection*{Generic detector oracle}

Like in~\cite{WHG:12}, we use the state-of-the-art Deformable Part Model
(DPM)~\cite{DPM} as our black box generic object detector.  This generic
detector is pre-trained on Pascal VOC 2007, and is publicly available
online\footnote{http://www.cs.berkeley.edu/~rbg/latent/voc-release5.tgz}.
%
%
The performance obtained by DPM is reported in Table~\ref{tab:detection}. 
%
In order to use DPM as a \emph{confident but laconic oracle}, we use only its top
$5\%$ detections.

\subsubsection*{Tracking-by-detection object detector}

Our formulation is applicable to any detector based on a linear classifier,
and is, therefore, usable with most state-of-the-art object detection and
window representation methods~\cite{DPM,Cinbis:13,girshick2014rcnn}.
In our experiments, we use Fisher Vectors (FV)~\cite{perronnin2007fisher}, as
this representation is particularly suited to our problem for the following
reasons.
First, FV are among state-of-the-art representations for object
detection~\cite{Cinbis:13}.
Second, they proved to be efficient for both category-level image
classification~\cite{Sanchez2013} and instance-level retrieval
problems~\cite{jegou2012pami}. This highlights their potential for both
category-level and instance-level appearance modeling.
Third, FV are high-dimensional representations, which allows our
mean-regularized multi-task learning to, in principle, be able to model complex
distributions using a single linear representation.
To the best of our knowledge, our method is the first application of FV to
tracking.

In our experiments, we followed the implementation of
Cinbis~\etal~\cite{Cinbis:13}, which yields competitive detection results on
Pascal VOC, with the difference that we use a more efficient but approximate
sliding window approach similar to the one described in
Oneata~\etal~\cite{Oneata2014}.

\subsection{Results}

\begin{table}
\begin{center}
\begin{tabular}{lcccccc}
                  & Ols1 & Ols2 & Olsr2 & Osow1 & Ose2 & VIRAT-0401 \\
\toprule
DPM~\cite{DPM}     & 30.4 & 52.4 & 34.9  & 52.2  & 34.8 &  47.0      \\
DbD~\cite{WHG:12} & {\bf 32.1} & 56.3 & 43.1  & 47.0  & {\bf 40.9} &  N.A.      \\
I-EIT             & 27.4 & 53.6 & 40.6  & 51.9  & 38.9 &  53.1      \\
{\bf EIT}   & 29.3 & {\bf 58.0} & {\bf 43.7}  & {\bf 53.1}  & 38.1 & {\bf 53.7} \\
\bottomrule
\end{tabular}
\end{center}
\caption{Detection performance (Average Precision). ``DPM'' is the pre-trained
generic detector (DPM~\cite{DPM}, using all detections, not just seeds). ``DbD''
is the re-ranking of the ``DPM'' detections using the algorithm of
Wang~\etal~\cite{WHG:12}. ``I-EIT'' is a baseline adapted detector from our
Ensemble of Instance Trackers without multi-task penalization, \ie with
independent trackers. ``EIT'' is the adapted detector obtained with our
``self-learning camera'' multi-task learning algorithm.}
\label{tab:detection}
\vspace*{-3mm}
\end{table}

Table~\ref{tab:detection} contains our quantitative experimental results.
Performance is measured using Average Precision (AP), which corresponds to
the area under the precision-recall curve. It is the standard metric used
to measure object detection performance~\cite{Everingham2010}.
Our results using the adapted detector obtained by our multi-task
Ensemble of Instance Trackers are reported in the ``EIT'' row.

We compare our approach to three other object detectors applicable in our
unsupervised setting: (i) the pre-trained DPM detector used alone over the whole video;
(ii) our implementation of the ``Detection by Detections'' (DbD) unsupervised
object detection approach of Wang~\etal~\cite{WHG:12}\footnote{``DbD'' relies on
re-ranking the DPM detections according to their similarities with the FV
descriptors of the highest scoring ones (the ones we use as seeds); note that
this is a batch off-line approach, therefore, not directly applicable to
scenarios like VIRAT};
(iii) our EIT algorithm without multi-task regularization, \ie
where the trackers are learned independently, which is referred to as
``I-EIT''.
Note that we do not compare to additional unsupervised video adaptation methods
(\eg~\cite{Tang:12,Sharma2012}), as our contribution is on autonomous on-line
self-adaptation, \ie we do not assume access to a large fixed labeled dataset
related to the target videos, or that all target video frames are available at
once (in contrast to~\cite{Tang:12,Sharma2012}).

First, we can observe that our approach improves over the generic detector
by $+4\%$ on average over the scenarios.
This confirms that unsupervised learning from scratch of our simple and
efficient detector for a specific video stream can outperform a state of the
art, carefully tuned, complex, and generic object detector trained on large
amounts of unrelated data.
Note that we improve in all scenarios but one (Ols1), which corresponds to
the smallest video, which is of roughly ten seconds (\cf
Table~\ref{tab:data}). This means that this video is not enough to learn a
detector that is better than the generic detector trained on much
more data (thousands of images in the case of Pascal VOC).

Second, jointly learning all trackers (EIT) improves by $+1.7\%$
on average over learning them independently. For some scenarios, the gain
is substantial ($+4.4\%$ on Ols2), while for one scenario it slightly
degrades performance ($-0.8\%$ on Ose2).
On the one hand, this illustrates that our multi-task tracking formulation can
help learning a better detector when the tracked seed instances share
appearance traits useful to recognize the category of objects of interest.
On the other hand, this calls for a more robust multi-task learning
algorithm than our simple mean-regularized formulation in order to handle
more robustly outliers and the large intra-class variations typically
present in broad object categories like cars.

Third, our approach yields a small improvement over the state-of-the-art
DbD method ($+0.6\%$ on average over the CAVIAR scenarios).
This suggests that our approach is accurate enough to discard the generic
detector after seeing a large enough part of the video stream (to get
enough seeds), whereas DbD must constantly run the generic detector, as it
is a ``lazy learning'' approach based on $k$-nearest neighbors (it does not
learn a stand-alone detector).
Note also that DbD could be applied as a post-processing step on the results of
our detector in order to improve performance, but this leads to additional
computational costs.


\bibliographystyle{ieeetr}
\bibliography{slc_references}

\end{document}